\documentclass[11pt]{article}
\usepackage{subfigure,bm} 
\usepackage[dvipdfmx]{graphicx}
\usepackage{authblk}

\usepackage[top=30truemm,bottom=30truemm,left=25truemm,right=35truemm]{geometry}

\usepackage{amssymb}
\usepackage{amsmath}

\DeclareMathOperator{\var}{\mathbf{Var}}

\newcommand{\T}{{\hspace{-0.25ex}\top\hspace{-0.25ex}}}
\newcommand{\bE}{\mathbb{E}}
\newcommand{\cN}{\mathcal{N}}
\newcommand{\cJ}{\mathcal{J}}

\title{\LARGE \bf
Efficient Reuse of Previous Experiences \\
 to Improve Policies in Real Environment
}

\author[1,3]{Norikazu Sugimoto}
\author[2]{Voot Tangkaratt}
\author[4]{Thijs Wensveen}
\author[2]{Tingting Zhao}
\author[2]{Masashi Sugiyama}
\author[1]{Jun Morimoto\thanks{xmorimo@atr,jp}}

\affil[1]{Dept. of Brain Robot Interface, ATR Computational Neuroscience Labs}
\affil[2]{Tokyo Institute of Technology}
\affil[3]{National Institute of Information and Communications Techonology}
\affil[4]{Delft University of Technology}

\begin{document}

\maketitle
\thispagestyle{empty}
\pagestyle{empty}

\begin{abstract}
In this study,
we show that a movement policy can be improved efficiently using the previous experiences of a real robot. 
Reinforcement Learning (RL) is becoming a popular approach to acquire a nonlinear optimal policy through trial and error.
However, it is considered very difficult to apply RL to real robot control since it usually requires many learning trials.
Such trials cannot be executed in real environments because unrealistic time is necessary 
and the real system's durability is limited. Therefore, in this study, instead of executing many learning trials, 
we propose to use a recently developed RL algorithm, importance-weighted PGPE, 
by which the robot can efficiently reuse previously sampled data to improve it's policy parameters. 
We apply importance-weighted PGPE to CB-i,
our real humanoid robot, and show that it can learn a target reaching
movement and a cart-pole swing up movement in a real environment 
without using any prior knowledge of the task or any carefully designed initial trajectory. 

\end{abstract}

\section{INTRODUCTION}

Reinforcement Learning (RL) is becoming a popular approach to acquire a nonlinear optimal policy through trial and error. 
However, it is considered very difficult to apply RL to real robot control since it usually requires many learning trials. 
Such learning trials cannot be executed in real environments because unrealistic time is necessary and  the real system's durability is limited.
 Therefore, previous studies used prior knowledge or properly designed initial trajectories to apply RL to a real robot so that the parameters 
 of a robot controller can be improved within a realistic amount of time \cite{Atkeson97ml,Matsubara06ras,IROS:Peters+Schaal:2006,HUMANOIDS:Sugimoto:2011}. 

However, since prior knowledge may not be always available, 
it is desirable to use a learning method that can improve policies only from limited experiences. 
In this study, we
consider a recently developed RL algorithm: importanceweighted
policy gradients with parameter based exploration
(IW-PGPE) [17]. With the IW-PGPE algorithm, we can
efficiently use previously sampled data to improve policies.
In other words, a robot can use its previous experiences to
improve the current policy parameters.
The usefulness of this approach has been thoroughly evaluated by comparing with previously proposed RL methods
\cite{Neurocomputing:Peters:2008,NECO:Hachiya:2011,ML:Williams:1992,ML:Sehnke:2010}
in numerical simulations. In this study, we evaluate how this RL approach using prevoius experiences can work efficiently in the real system.

\begin{figure}[t]
 \begin{center}
  \includegraphics[width=70mm]{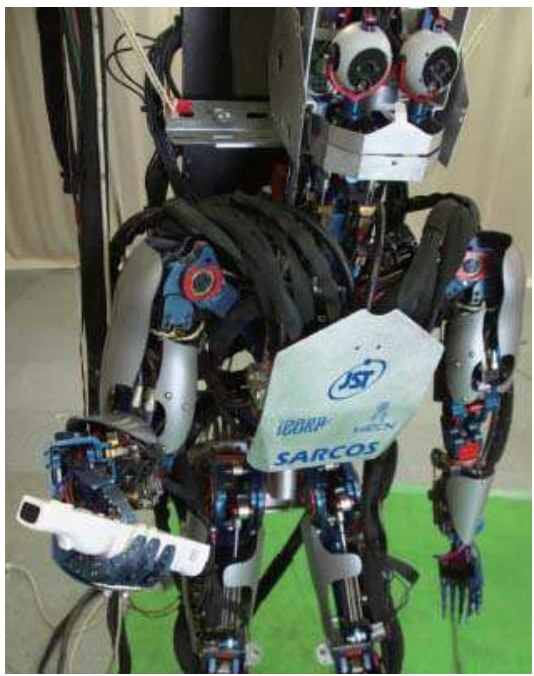}
  \caption{Humanoid robot CB-i \cite{cbi} grabbing Wii controller.}
  \label{fig:cbi}
 \end{center}
\end{figure}

We apply IW-PGPE to our real humanoid robot called CB-i \cite{cbi} (see Fig. \ref{fig:cbi})
and show that it can learn a target-reaching movement
and a cart-pole swing-up movement in a real environment
within 1.5 hours without using any prior knowledge of the task or any initial trajectory.
For a target-reaching task, we  used five degrees of freedom (DOF) composed of one-DOF torso joint, three-DOF shoulder joints,
and one-DOF elbow joint of the humanoid robot.  To use these five DOFs,
the policy needs to be improved in ten-dimensional state space. 
In this moderately high-dimensional state space, it is usually considered that RL cannot be directly applied to real systems.
However, we show that a target-reaching policy can be improved within a realistic amount of time by using IW-PGPE. 

The rest of this paper is organized as follows. Policy update methods by using the standard policy gradient method and by using  PGPE 
are explained in Section \ref{sec:PolicyGradientMethods}.
The extension of PGPE to efficiently use previous experiences is introduced in Section 
\ref{sec:EfficientReuseOfPrefiousExperiences}.
In Section \ref{sec:ExperimentalResults},  we evaluate our approach to improve policies in a real environment.
First, we show the performances of policy updates by the IW-PGPE method
with different parameters in a simulation environment.
Then, we apply the learning method to our humanoid robot CB-i to improve a target-reaching policy.
In section \ref{sec:ExperimentalResults2},
we introduce a newly developed experimental setup
in which our real humanoid robot can interact with a virtual environment through a Wii controller.
In this setup, a cart-pole swing-up policy is learned.
Finally, Section \ref{sec:Conclusion} concludes the paper with discussion and following work in the future.

\section{Policy Gradient Methods}
\label{sec:PolicyGradientMethods}
In this section, we first review a standard formulation of policy gradient methods
\cite{mach:Williams:1992,nips02-CN11,IROS:Peters+Schaal:2006}.
Then we show an alternative formulation adopted in the PGPE (policy gradients with
parameter based exploration) method \cite{NN:Sehnke+etal:2010}.

\subsection{Standard Policy Update}
\label{subsec:RLformulation}
We assume that the underlying control problem is a discrete-time MDP.
At each discrete time step $t$, the agent observes a state $\bm{x}_t \in {\cal X}$, selects an action $u_t \in {\cal U}$,
and then receives an immediate reward $r_t$ resulting from a state transition in the environment.
The dynamics of the environment are characterized by $p(\bm{x}_{t+1}|\bm{x}_{t},u_t)$, which represents the transition probability density from the current state $\bm{x}_t$ to
the next state $\bm{x}_{t+1}$ when action $u_t$ is taken, and $p(\bm{x}_1)$ is the probability density of initial states.
The immediate reward $r_t$ is given according to the reward function $r(\bm{x}_t,u_t,\bm{x}_{t+1})$.

The robot's decision making procedure at each time step $t$ is characterized by a parameterized policy $p(u_t|\bm{x}_t,\bm{\theta})$
with parameter $\bm{\theta}$,
which represents the conditional probability
density of taking action $u_t$ in state $\bm{x}_t$.
We assume that the policy is continuously differentiable with respect to its parameter $\bm{\theta}$.

A sequence of states and actions forms a \emph{trajectory} denoted by
\[
h:=[\bm{x}_{1},u_{1},\ldots,\bm{x}_{T},u_{T}],
\]
where $T$ denotes the number of steps called horizon length.
In this paper, we assume that $T$ is a fixed deterministic number.
Note that the action $u_t$ is chosen independently of the trajectory given $\bm{x}_t$ and $\bm{\theta}$.
Then the discounted cumulative reward along $h$, called the \emph{return},
is given by
\[
R(h):=
\sum_{t=1}^T \gamma ^{t-1}r(\bm{x}_t,u_t,\bm{x}_{t+1}),
\]
where $\gamma \in [0, 1)$ is the discount factor for future rewards.

The goal is to optimize the policy parameter $\bm{\theta}$ so that the
\emph{expected return} is maximized.
The expected return for policy parameter $\bm{\theta}$ is defined by
\[
J(\bm{\theta}):=\int p(h|\bm{\theta})R(h)\mathrm{d}h, \]
where
\[ p(h|\bm{\theta})=p(\bm{x}_1)\prod_{t=1} ^T p(\bm{x}_{t+1}|\bm{x}_t,u_t)p(u_t|\bm{x}_t,\bm{\theta}).\]
The most straightforward way to update the policy parameter is to follow the gradient in policy parameter space using gradient ascent:
\[
\bm{\theta}\longleftarrow\bm{\theta}+\varepsilon\nabla_{\bm{\theta}} J(\bm{\theta}),
\]
where $\varepsilon$ is a small positive constant, called the learning rate.

This is a standard formulation of policy gradient methods \cite{mach:Williams:1992,nips02-CN11,IROS:Peters+Schaal:2006}.
The central problem is to estimate the policy gradient $\nabla_{\bm{\theta}} J(\bm{\theta})$ accurately from trajectory samples.

\subsection{PGPE Policy Update}

However, standard policy gradient methods were shown to suffer from high variance in the gradient estimation
due to randomness introduced by the stochastic policy model $p(a|\bm{s},\bm{\theta})$ \cite{NN2012-ting}.
To cope with this problem, an alternative method
called \emph{policy gradients with parameter based exploration} (PGPE) was proposed
recently \cite{NN:Sehnke+etal:2010}.
The basic idea of PGPE is to use a deterministic policy
and introduce stochasticity by drawing parameters from a prior distribution.
More specifically, parameters are sampled from the prior distribution at the start of each trajectory,
and thereafter the controller is deterministic\footnote{Note that transitions are stochastic, and thus trajectories are also stochastic even though the policy is deterministic.}.
Thanks to this per-trajectory formulation,
the variance of gradient estimates in PGPE does not increase with respect to trajectory length $T$.
Below, we review PGPE.

PGPE uses a deterministic policy with typically a linear architecture:
\begin{align}
p(u|\bm{x},\bm{\theta})
=\delta(u=\bm{\theta}^{\top} \bm{\phi} (\bm{x})),
\label{linear-policy-model}
\end{align}
where $\delta(\cdot)$ is the \emph{Dirac delta function},
$\bm{\phi} (\bm{s})$ is an $\ell$-dimensional basis function vector,
and $^\T$ denotes the transpose.
The policy parameter $\bm{\theta}$ is drawn from
a prior distribution $p(\bm{\theta}|\bm{\rho})$ with hyper-parameter $\bm{\rho}$.

The expected return in the PGPE formulation is defined
in terms of expectations over both $h$ and $\bm{\theta}$
as a function of hyper-parameter $\bm{\rho}$:
\[
\cJ(\bm{\rho}):=\iint p(h|\bm{\theta})p(\bm{\theta}|\bm{\rho})
R(h)\mathrm{d}h\mathrm{d}\bm{\theta}.
\]
In PGPE, the hyper-parameter $\bm{\rho}$ is optimized so as to maximize $\cJ(\bm{\rho})$,
i.e., the optimal hyper-parameter $\bm{\rho}^*$ is given by
\[
\bm{\rho}^*:= \arg\max_{\bm{\rho}} \cJ(\bm{\rho}).
\]

In practice, a gradient method is used to find $\bm{\rho}^*$:
\[
\bm{\rho}\longleftarrow\bm{\rho}+\varepsilon\nabla_{\bm{\rho}} \cJ(\bm{\rho}),
\]
where
$\nabla_{\bm{\rho}} \cJ(\bm{\rho})$ is the derivative of $\cJ$ with respect to $\bm{\rho}$:
\begin{align}
\nabla_{\bm{\rho}} \cJ(\bm{\rho})=\iint p(h|\bm{\theta})p(\bm{\theta}|\bm{\rho})  \nabla_{\bm{\rho}} \log p(\bm{\theta}|\bm{\rho}) R(h)\mathrm{d}h\mathrm{d}\bm{\theta}.
\end{align}
Note that, in the derivation of the gradient, the logarithmic derivative,
\[
\nabla_{\bm{\rho}}  \log p(\bm{\theta}|\bm{\rho})=
\frac{\nabla_{\bm{\rho}}p(\bm{\theta}|\bm{\rho})}{p(\bm{\theta}|\bm{\rho})},
\]
was used.
The expectations over $h$ and $\bm{\theta}$ are approximated by the empirical averages:
\begin{align}
\label{emp_gra}
\nabla_{\bm{\rho}} \widehat{\cJ}(\bm{\rho}) = \frac {1} {N} \sum ^N_{n=1} \nabla_{\bm{\rho}} \log p(\bm{\theta}_n|\bm{\rho})R(h_n),
\end{align}
where each trajectory sample $h_n$ is drawn independently from $p(h|\bm{\theta}_n)$ and
parameter $\bm{\theta}_n$ is drawn from $p(\bm{\theta}_n|\bm{\rho})$.
We denote samples collected at the current iteration as
\[
D=\{\left(\bm{\theta}_n, h_n\right)\}_{n=1}^N.
\]

Following \cite{NN:Sehnke+etal:2010}, in this paper we employ a Gaussian distribution as the distribution of the policy parameter $\bm{\theta}$
with the hyper-parameter $\bm{\rho}$.
When assuming a Gaussian distribution,
the hyper-parameter $\bm{\rho}$ consists of a set of means $\{{\eta}_i\}$ and standard deviations $\{\tau_i\}$,
which determine the prior distribution for each element $\theta_i$ in $\bm{\theta}$ of the form
\[p(\theta_i|\rho_i)=\cN(\theta_i|\eta_i,\tau_i^2),\]
where $\cN(\theta_i|\eta_i,\tau_i^2)$ denotes the normal distribution with mean $\eta_i$ and variance $\tau_i^2$.
Then the derivative of $\log p(\bm{\theta}|\bm{\rho})$ with respect to $\eta_i$ and $\tau_i$ are given as
\begin{align*}
  \nabla_{\eta_i}\log p(\bm{\theta}|\bm{\rho})=&\frac{\theta_i-\eta_i}{\tau_i^2},\\
  \nabla_{\tau_i} \log p(\bm{\theta}|\bm{\rho})=&\frac{(\theta_i-\eta_i)^2-\tau_i^2}{\tau_i^3},
\end{align*}
which can be substituted into Eq.(\ref{emp_gra}) to approximate the
gradients with respect to $\bm{\eta}$ and $\bm{\tau}$.
These gradients give the PGPE update rules.

An advantage of PGPE is its low variance of gradient estimates:
Compared with a standard policy gradient method REINFORCE \cite{mach:Williams:1992},
PGPE was empirically demonstrated to be better in some settings \cite{NN:Sehnke+etal:2010, NN2012-ting}.
The variance of gradient estimates in PGPE
can be further reduced by subtracting an optimal baseline. 


\section{Efficient Reuse of Previous Experiences}
\label{sec:EfficientReuseOfPrefiousExperiences}

The original PGPE is categorized as an \emph{on-policy} algorithm \cite{book:Sutton+Barto:1998},
where data drawn from the current target policy is used to estimate policy gradients.
On the other hand, \emph{off-policy} algorithms are more flexible in the sense
that a data-collecting policy and the current target policy can be different.
In this section, we introduce an \emph{off-policy} algorithm for PGPE  proposed by  \cite{NECO:Tingting:2013}.
In this algorithm, importance-weighting is used so that
we can reuse previously collected data (experience) in a consistent manner.

\subsection{Importance-Weighted PGPE}

Let us consider an off-policy scenario where
a data-collecting policy and the current target policy are different in general.
In the context of PGPE, we consider two hyper-parameters, $\bm{\rho}$ for the target policy to learn
and $\bm{\rho}'$ for data collection. Let us denote data samples collected with hyper-parameter $\bm{\rho}'$ by $D'$:
\[
D'=\{\left(\bm{\theta}'_n, h'_n\right)\}_{n=1}^{N'} \overset{i.i.d}{\sim}
p(h,\bm{\theta}|\bm{\rho}')=p(h|\bm{\theta})p(\bm{\theta}|\bm{\rho}').
\]
If we naively use data $D'$ to estimate policy gradients by Eq.\eqref{emp_gra},
we have an inconsistency problem:
\[
\frac {1} {N'} \sum ^{N'}_{n=1} \nabla_{\bm{\rho}} \log p(\bm{\theta}'_n|\bm{\rho})R(h_n')
\overset{N'\rightarrow\infty}{\nrightarrow}
\nabla_{\bm{\rho}} \cJ(\bm{\rho}).
\]

\emph{Importance sampling} \cite{book:Fishman:1996}
is a technique to systematically resolve this distribution mismatch problem.
The basic idea of importance sampling is to weight samples drawn from a sampling distribution
to match the target distribution, which gives a consistent gradient estimator:
\begin{eqnarray}
&\nabla_{\bm{\rho}}\widehat{\cJ}_{\mathrm{IW}}(\bm{\rho})
:=\frac {1} {N'}\sum ^{N'}_{n=1} w(\bm{\theta}'_n) \nabla_{\bm{\rho}} \log p(\bm{\theta}'_n|\bm{\rho})R(h_n') \nonumber \\
&\hspace{50mm}\overset{N'\rightarrow\infty}{\longrightarrow} \nabla_{\bm{\rho}} \cJ(\bm{\rho}),
\end{eqnarray}
where
\[
w(\bm{\theta})=\frac{p(\bm{\theta}|\bm{\rho})}{p(\bm{\theta}|\bm{\rho}')}
\]
is called the \emph{importance weight}.

An intuition behind importance sampling is that if we know how ``important''
a sample drawn from the sampling distribution is in the target distribution,
we can make adjustment by importance weighting.
This extended method is called \emph{importance-weighted PGPE} (IW-PGPE) \cite{NECO:Tingting:2013}.

\subsection{Variance Reduction by Baseline Subtraction for IW-PGPE}
\label{sec:variance}


To further reduce the variance of gradient estimates in IW-PGPE,
we use variance reduction technique which uses
a constant \emph{baseline} \cite{sutton:dissertation84,Williams:88,JMLR:Greensmith+Bartlett+Baxter:2004,UAI:Weaver+Tao:2001}
as suggested in \cite{NECO:Tingting:2013}.


A policy gradient estimator with a baseline $b\in\mathbb{R}$ is defined as
\[
\nabla_{\bm{\rho}}\widehat{\cJ}^b_{\mathrm{IW}}(\bm{\rho})
:=\frac{1}{N'}\sum_{n=1}^{N'} (R(h'_n)-b) w(\bm{\theta}'_n) \nabla_{\bm{\rho}} \log p(\bm{\theta}'_n|\bm{\rho}).
\]
 It is well known that $\nabla_{\bm{\rho}}\widehat{\cJ}^b_{\mathrm{IW}}(\bm{\rho})$
is still a consistent estimator of the true gradient for any constant $b$ \cite{JMLR:Greensmith+Bartlett+Baxter:2004}.
Here, the constant baseline $b$ is determined so that the variance is minimized.
Let $b^*$ be the optimal constant baseline for IW-PGPE that minimizes the variance:
\[
b^*:= \arg\min_b \var[\nabla_{\bm{\rho}}\widehat{\cJ}_{\mathrm{IW}}^b(\bm{\rho})].
\]

\label{theorem:optimal-baseline}
The optimal constant baseline for IW-PGPE is derived as \cite{NECO:Tingting:2013}:
\[
b^*= \frac{\bE_{p(h,\bm{\theta}|\bm{\rho'})}[R(h)w^2(\bm{\theta})\| \nabla_{\bm{\rho}} \log p(\bm{\theta}|\bm{\rho})\|^2]}
      {\bE_{p(h,\bm{\theta}|\bm{\rho'})}[w^2(\bm{\theta})\|\nabla_{\bm{\rho}} \log p(\bm{\theta}|\bm{\rho})\|^2]}.\]

\begin{figure}[t]
 \begin{center}
  \includegraphics[width=60mm]{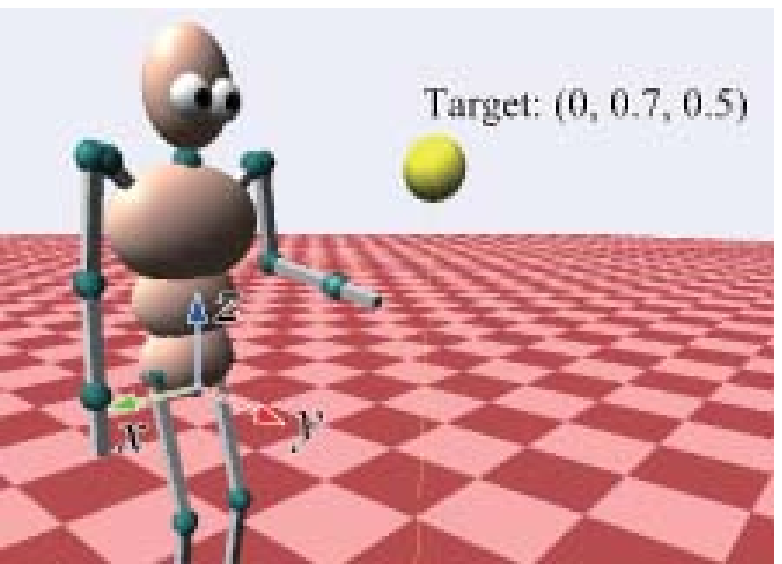}
  \caption{Humanoid robot simulator ``SL'' \cite{TheSL_Simulation_and_RealtimeControlSoftwarePackage}.}
  \label{fig:xsugi_SL}
 \end{center}
  \begin{center}
  \includegraphics[width=60mm]{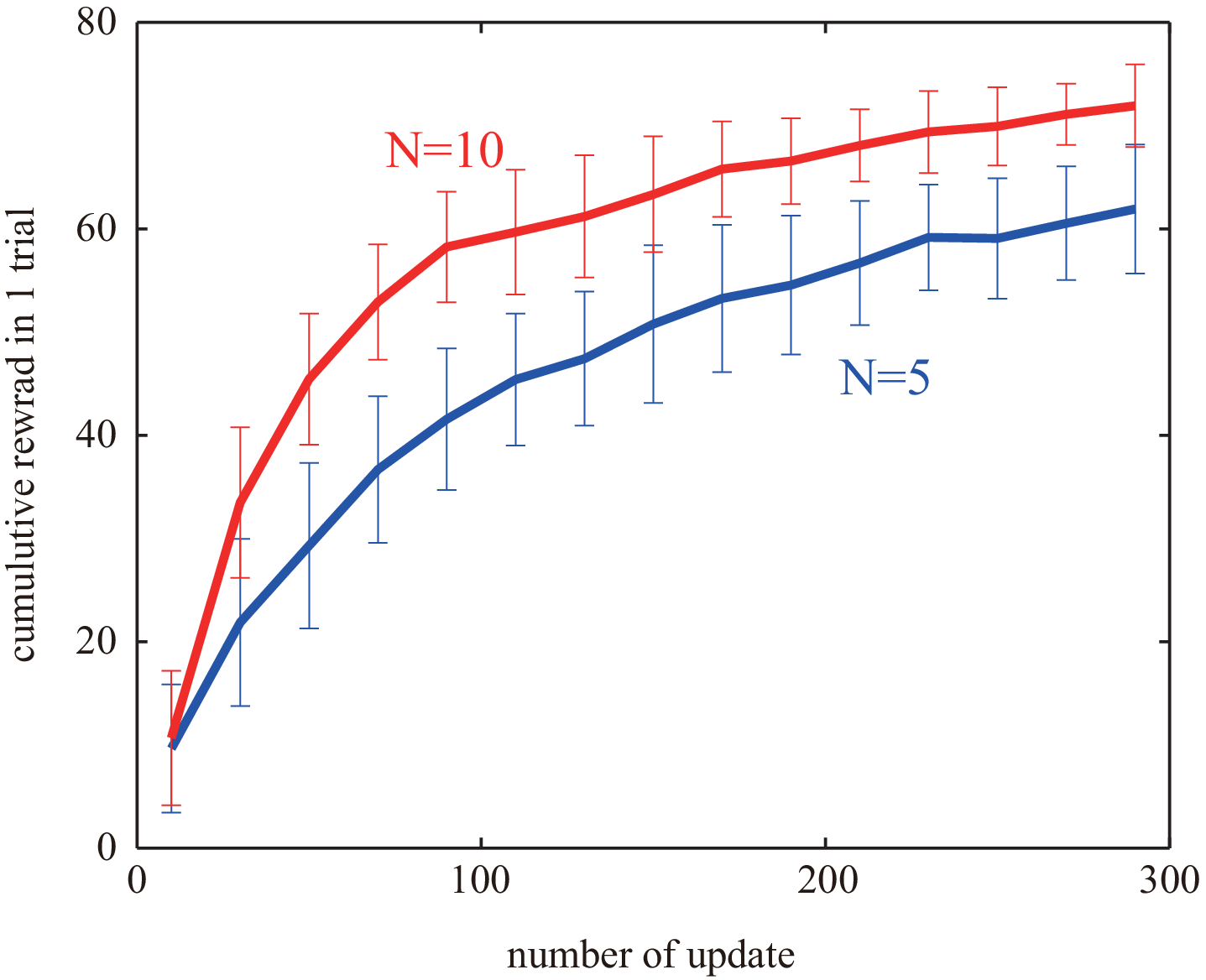}
  \caption{Learning curve of cumulative reward. The horizontal and vertical axes are number of update and cumulative reward.
  Blue and red lines represent the case that parameters were updated every five trials ($N=5$) and  ten trials ($N=10$), respectively.}
  \label{fig:result_sim_EPI05_EPI10}
 \end{center}
\end{figure}


\section{Experimental Results: Hand reaching task}
\label{sec:ExperimentalResults}

We applied our proposed approach to a target-reaching task in simulated and real environments.
Fig. \ref{fig:xsugi_SL} shows a setup of the target-reaching task.
The end-effector is left-hand of the humanoid robot and the target is placed in front of the robot.

In this task, the robot controls 5 joints of the upper body, yaw joint of torso, three joints of left-shoulder, and joint of left-elbow,
and learns policy to reach the target.
The length of one trial was $1$ sec.

The controller outputs a desired velocities of each joint:
\begin{align}
 \dot{\psi}^{\rm des}_i(t) &= \bm{\theta}^{\top} \bm{\phi} (\bm{x}(t)),
\end{align}
where $\psi^{\rm des}_i (i = \{1, 2, 3, 4, 5\})$ are the desired trajectory of controlled joint.
In this paper, we employed linear basis function:
\begin{align}
 \bm{\phi}(\bm{x}(t)) &= \left[\psi_1(t) \,\,\,\, \cdots \,\,\,\, \psi_5(t) \,\,\,\, \dot{\psi}_1(t) \,\,\,\, \cdots \,\,\,\, \dot{\psi}_5(t) \,\,\,\, 1 \right]^{\T},
\end{align}
where $\psi_i (i = \{1, 2, 3, 4, 5\})$ are the actual joint angle.
The PD controller outputs the torque command $\tau_i$ for each joint to track the desired trajectories,
\begin{align}
 \tau_i &= - K_{\rm P} (\psi_i - \psi^{\rm des}_i) - K_{\rm D} (\dot{\psi}_i - \dot{\psi}^{\rm des}_i),
\end{align}
where $K_{\rm P}$ and $K_{\rm D}$ are positive constant.

We defined the objective function as sum of a state dependent reward $q(\bm{x}(t))$
and a cost of an action $c(\bm{x}(t), \bm{u}(t))$:
\begin{align}
 r(\bm{x}(t),\bm{u}(t)) &= q(\bm{x}(t)) - c(\bm{x}(t), \bm{u}(t)).
\end{align}
A state dependent reward is given based on the error between the end-effector and targets:
\begin{align}
 q(\bm{x}(t)) &= \exp \big[ -\alpha \| \bm{p}_{\mathbf E}(t) - \bm{p}_{\mathbf T} \|^2\big],
\end{align}
where $\bm{p}_{\mathbf E}(t)$ and $\bm{p}_{\mathbf T} = (0.5, 0.7, 0)$
are the end-effector position and the target position.
The origin is the center of torso joints.
The parameter $\alpha(=10)$ is constant.
The cost of control is given based on the difference between actual angle and desired angle,
\begin{align}
 c(\bm{x}(t), \bm{u}(t)) &= \beta \sum^{5}_{i=1} (\psi_i(t) - \psi^{\rm des}_i(t))^2,
\end{align}
where $\beta(=0.0005)$ is constant.
To maximize the future cumulative reward,
we updated the parameter $\bm{\theta}$
with the discount factor $\gamma=0.999$ and the learning rate $\varepsilon=0.1$.

\subsection{Simulation}
We first apply the proposed approach to the simulated environment.
We evaluate the learning performance of the proposed approach
with the different settings of  parameters of the learning algorithm, $N$ and $N'$,
We tested proposed method using humanoid robot simulator ``SL''
\cite{TheSL_Simulation_and_RealtimeControlSoftwarePackage} (See Fig.\ref{fig:xsugi_SL}).
To evaluate average learning performances of the proposed approach with different parameter settings,
we simulated five runs with different random seeds.

Figure \ref{fig:result_sim_EPI05_EPI10} shows the result of learning.
The horizontal and vertical axes are number of update and cumulative reward.
Two lines (blue and red) represent the performance for $N=5$ and $N'=10$.

\begin{figure}[t]
 \begin{center}
  \includegraphics[width=60mm]{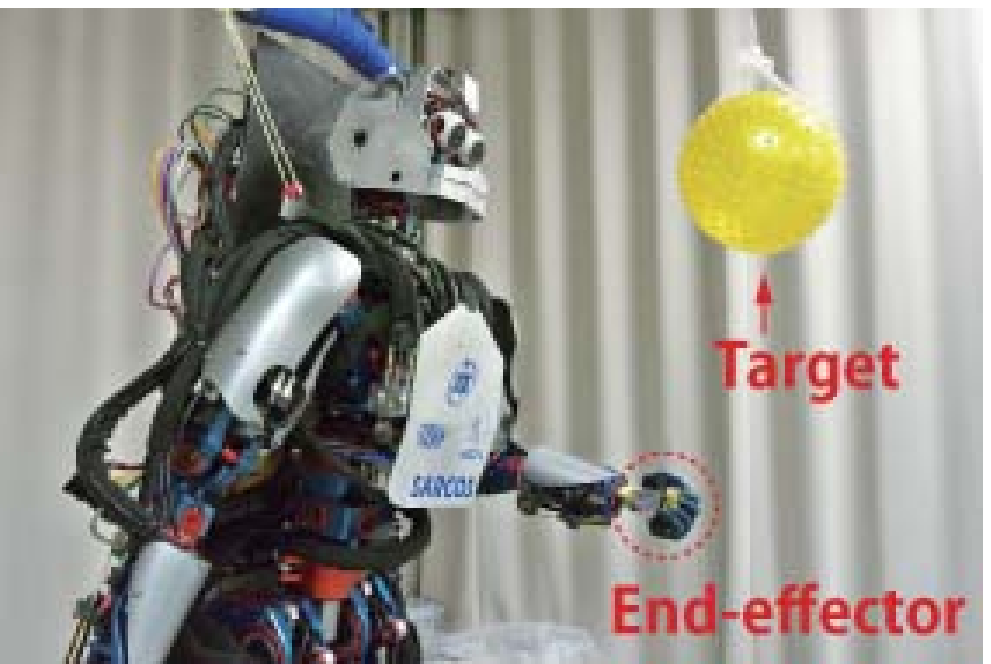}
  \caption{Humanoid robot ``CB-i''.
  The end-effector of reaching is right hand. The ball represent the target of reaching.}
  \label{fig:xsugi_CB-i}
 \end{center}
  \begin{center}
  \includegraphics[width=60mm]{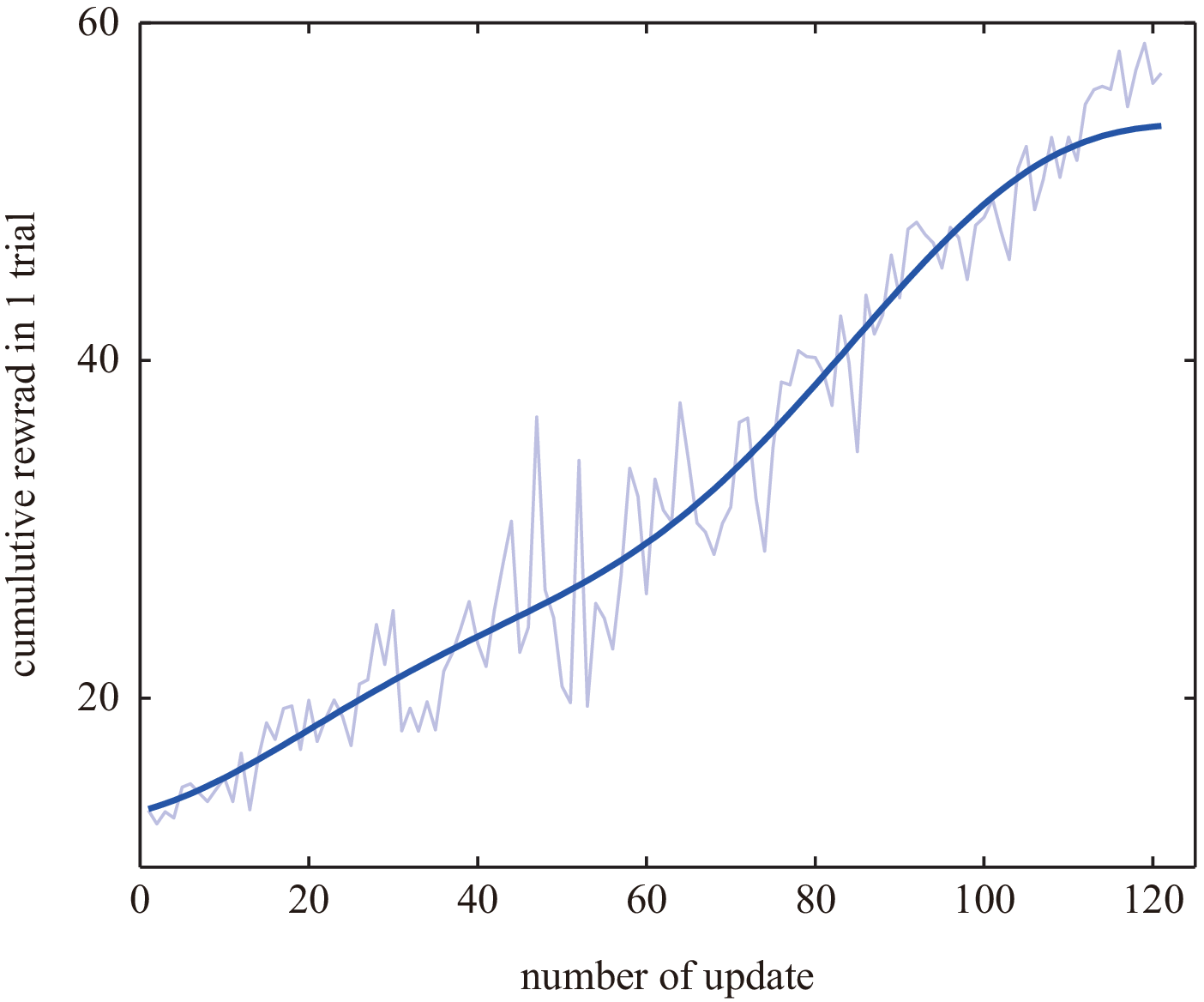}
  \caption{Learning curve of cumulative reward in the real humanoid robot experiment.
  The horizontal and vertical axes are number of update and cumulative reward.}
  \label{fig:result_exp}
 \end{center}
\end{figure}


\subsection{Real humanoid robot experiment}
Finally, we implement proposed approach to real humanoid robot ``CB-i'' (See Fig. \ref{fig:xsugi_CB-i}).

In this experiment, we updated the parameters $\bm{\theta}$ every five trial and used $10$ trials for policy update ($N=5$ and $N'=10$).
The experiment's conditions are same as simulation except for the hyper parameters $N$ and $N'$.

\begin{figure}[!t]
 \begin{center}
  \includegraphics[width=65mm]{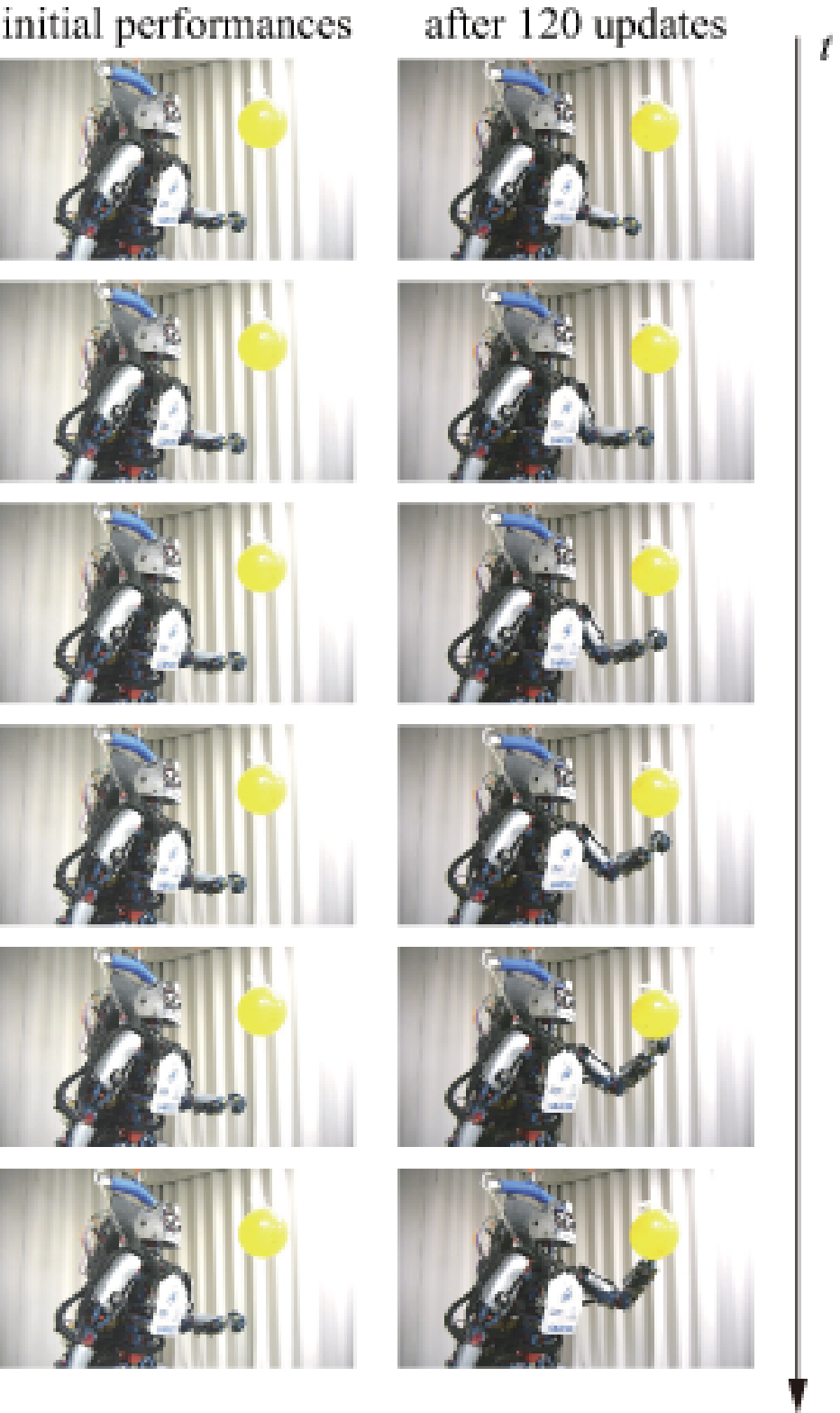} 
  \caption{Acquired behavior of humanoid robot.}
  \label{fig:xsugi_cbi_strove}
 \end{center}
\end{figure}

Figure \ref{fig:result_exp} shows the learning curve of cumulative reward.
The horizontal and vertical axes are number of update and cumulative reward.
After 120 iterations, the end-effector reached target perfectly.
Fig. \ref{fig:xsugi_cbi_strove} shows acquired behavior at 120th iteration.

\begin{figure}[!t]
 \begin{center}
  \includegraphics[width=85mm]{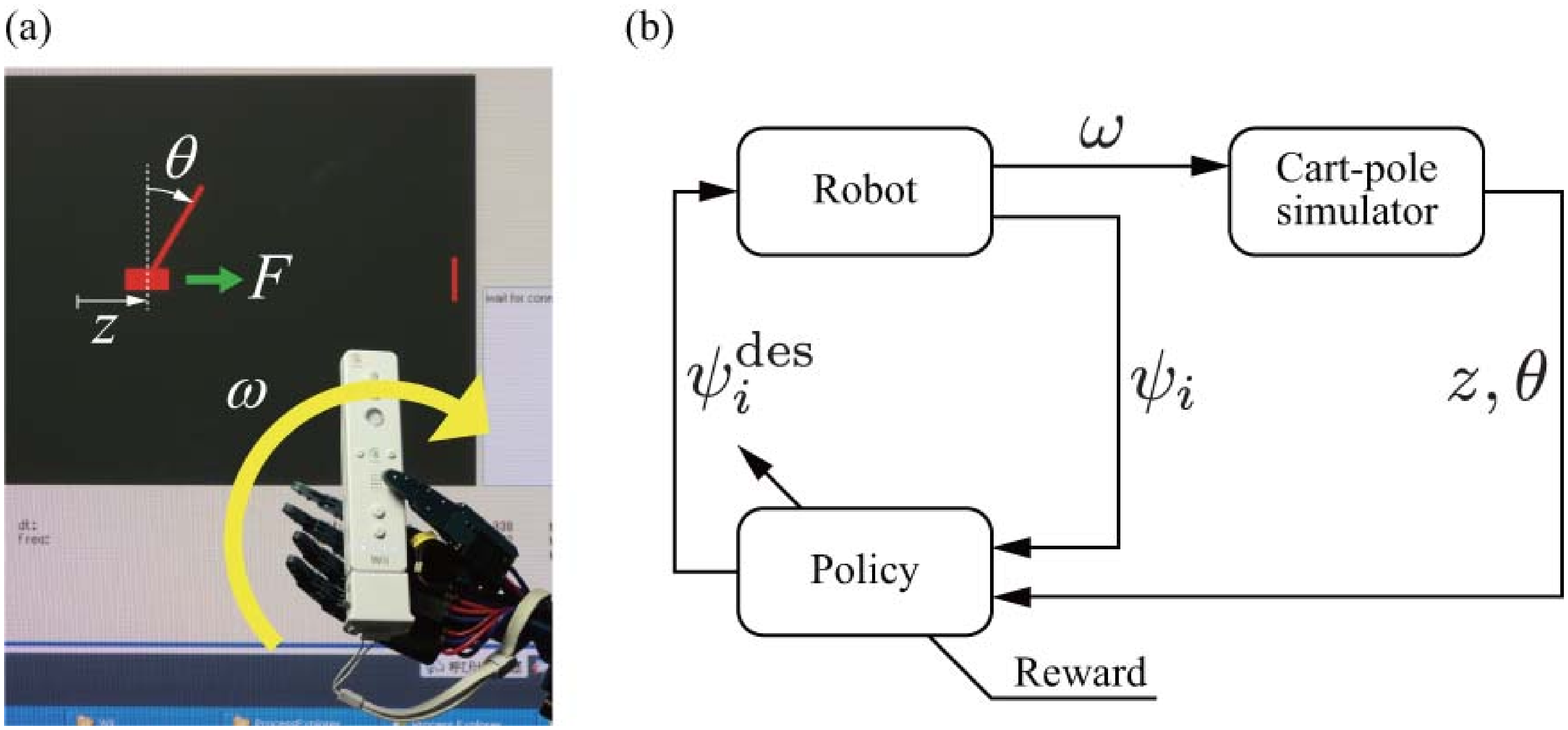} 
  \caption{(a) Robot controls a cart-pole which is simulated in the virtual environment to swing-up from hanging position.
  The angular velocity $\omega$ is converted to the driving force of the cart.
  (b) A diagram of experimental system.
  The desired trajectory of controlled joint ($\psi^{\rm des}_i$) is given to the robot,
  then the angular velocity of wii-controller is given to the cart-pole simulator.
  The actual trajectory of the robot and the cart-pole ($\psi_i$, $z$, $\theta$) is feed-backed to the policy.
  The policy is optimized to maximize the cumulative reward.}
  \label{fig:xsugi_exp_wii_cartpole}
 \end{center}
  \begin{center}
  \includegraphics[width=60mm]{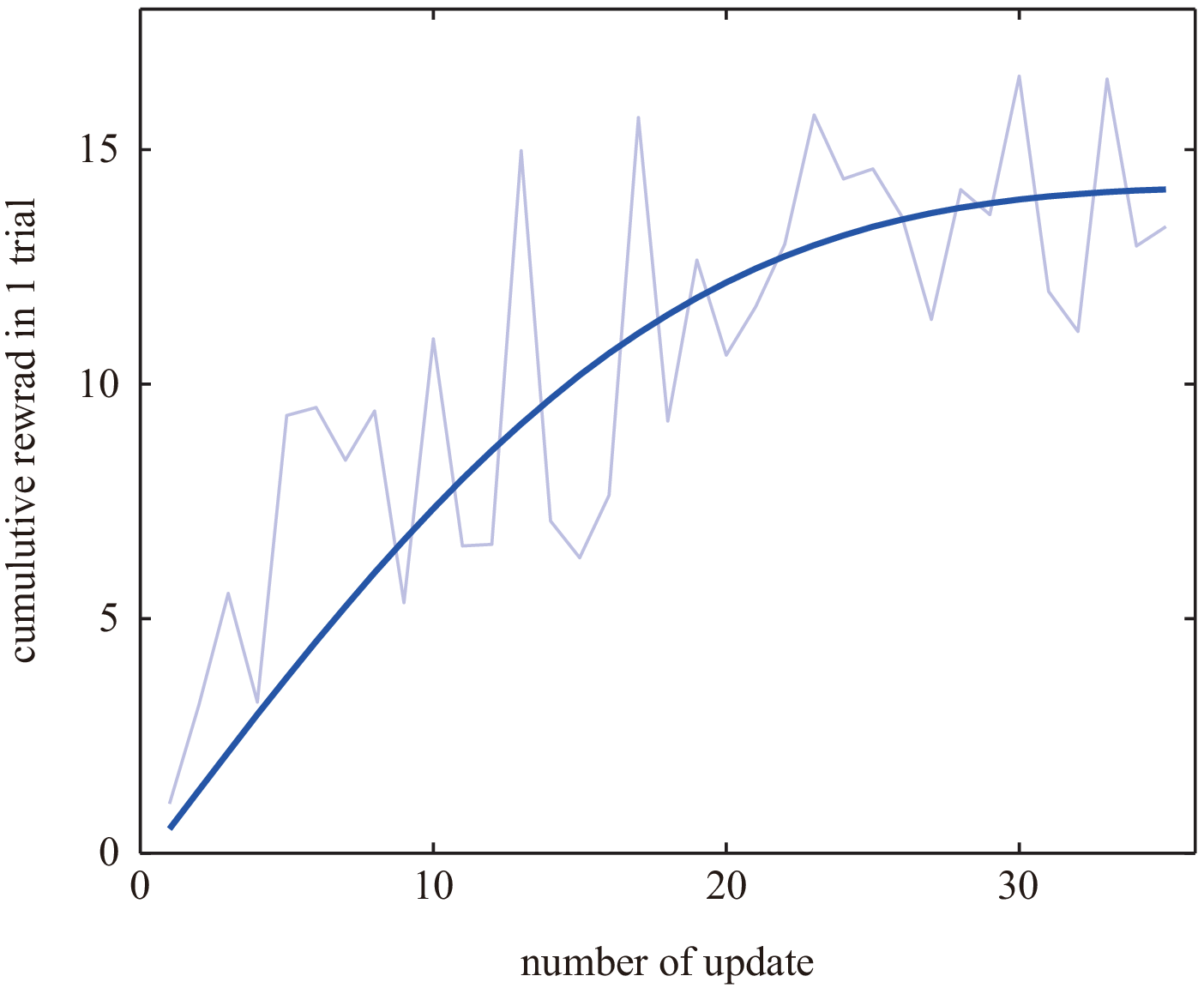}
  \caption{Learning curve of cumulative reward in the cart-pole swing-up task using the real humanoid robot.}
  \label{fig:xsugi_exp_wii_cum_reward}
 \end{center}
\end{figure}


\begin{figure}[t]
 \begin{center}
  \includegraphics[width=65mm]{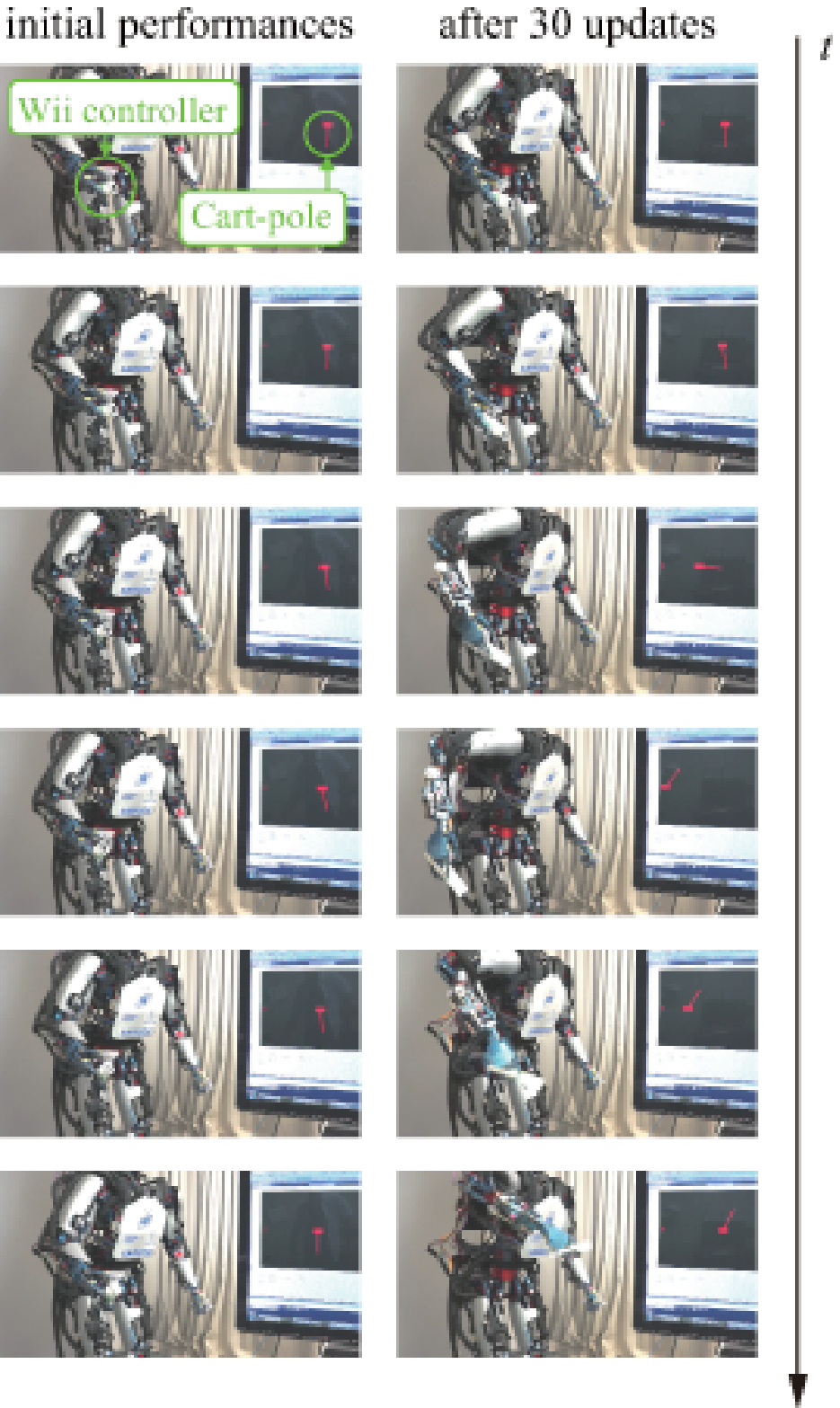} 
  \caption{Acquired behavior of humanoid robot in the cart-pole swing-up task.}
  \label{fig:xsugi_exp_wii_strove}
 \end{center}
\end{figure}

\section{Experimental Result: Cart-pole swing-up task}
\label{sec:ExperimentalResults2}
Next, we tested more challenging motor-control task using a virtual environment.
As see Fig.\ref{fig:xsugi_exp_wii_cartpole}, we developed a virtual dynamics simulator and
the robot interact with the virtual environment using a Wii controller.
In this task, the robot controls a cart-pole to swing-up from hanging position by swinging a Wii controller.
The dynamics of cart-pole is under actuated: the robot can apply a force to the cart only.

In this task, the robot also controls five joints of the upper body.
The cart-pole has two joints: horizontal position of the cart and angle of the pole.
The length of one trial was 1 sec.

We also employed linear basis function:
\begin{align}
 \bm{\phi}(\bm{x}(t)) &= \left[ \bm{s}(t)^{\T} \,\,\,\, \dot{\bm{s}}(t)^{\T} \,\,\,\, 1 \right]^{\T}, \\
 \bm{s} &= \left[ \psi_1(t) \,\,\,\, \cdots \,\,\,\, \psi_5(t) \,\,\,\, z(t) \,\,\,\, \theta(t) \right]^{\T},
\end{align}
where $z$ and $\theta$ are horizontal position of the cart and angle of the pole.

The controlling force for the cart is generated based on an angular velocity of wii-controller:
\begin{align}
 F(t) &= -k(\dot{z}(t) - \alpha \omega), \label{eq:exp_cart_force}
\end{align}
where $k=100$ and $\alpha=5$ are constant parameters.

We also defined the objective function as sum of a state dependent reward $q(\bm{x}(t))$
and a cost of an action $c(\bm{x}(t), \bm{u}(t))$.
A state dependent reward is given based on the angle of the pole:
\begin{align}
 q(\bm{x}(t)) &= \exp \big[ -\alpha (z(t)^2 + \theta(t)^2) \big],
\end{align}
where the parameter $\alpha=1$ is a constant.
When the position of the cart is center and the angle of pole is upright position ($z=0$, $\theta=0$), the state dependent reward becomes maximum value.
The cost of control is given based on the difference between actual angle and desired angle,
\begin{align}
 c(\bm{x}(t), \bm{u}(t)) &= \beta \sum^{5}_{i=1} (\psi_i(t) - \psi^{\rm des}_i(t))^2,
\end{align}
where $\beta(=0.0005)$ is a constant.
To maximize the future cumulative reward,
we updated the parameter $\bm{\theta}$
with the discount factor $\gamma=0.999$ and the learning rate $\varepsilon=0.1$.
The parameters of policy were updated every five trials ($N=5$) with sample size $N'=10$.
Note that we only consider swing-up movement and do not consider stabilizing the pole at upright position
because a Wii controller is not accurate enough to do the pole stabilizing.


\subsection{Real humanoid robot experiment}
Figure \ref{fig:xsugi_exp_wii_cum_reward} shows the learning result of the experiment.
The horizontal and vertical axes are the number of iteration and the mean cumulative reward in each iteration respectively.
The performance was reached maximum value around $20^{\rm th}$ iteration.
The learning speed was faster than the case of reaching task.
It was because that the parameter of reward function ($\alpha$) was different.
Fig. \ref{fig:xsugi_exp_wii_strove} shows the initial and acquired swing-up movements of the cart-pole at 30th iteration.


\section{CONCLUSIONS}
 \label{sec:Conclusion} 
In this study,
we show that the target-reaching policies can be efficiently acquired
by using the previous experiences of the robot in the real environment.
To improve the target-reaching policy, 
we used recently proposed IW-PGPE algorithm \cite{NECO:Tingting:2013}.
We also evaluated the learning performance  
with different parameter settings.
Moreover, we introduced newly developed real-virtual hybrid experimental setup
and showed that the real humanoid were able to swing up the virtual pole by using Wii controller.
As a future study, we will consider to develop a method to 
find the appropriate number of previously collected data that is used
to improve the current policy.






\section*{ACKNOWLEDGMENT}
This research was supported by MEXT KAKENHI 23120004.



\bibliographystyle{plain}

\end{document}